\documentclass[conference, a4paper]{IEEEtran}
\IEEEoverridecommandlockouts

\usepackage{tikz}
\usetikzlibrary{shapes.geometric, arrows}
\usetikzlibrary{automata,positioning,
shadows,
shapes.multipart,shapes.misc,backgrounds} 
\tikzset{every state/.style={minimum size=10pt } }
\usetikzlibrary{backgrounds,fit,decorations.pathreplacing}  
\usepackage{cite}
\usepackage{wrapfig}
\usepackage{subcaption}
\usepackage{amsmath,amssymb,amsfonts}
\usepackage{algorithmic}
\usepackage{algorithm}
\usepackage{graphicx}
\usepackage{graphics}
\usepackage{textcomp}
\usepackage{xcolor}
\usepackage{tabularx}
\usepackage{makecell, multirow}
\usepackage{float}
\usepackage{caption}
\usepackage{subcaption}
\usepackage{notes2bib}
\usepackage{colortbl}
\usepackage[T1]{fontenc}
\usepackage{float}

\usepackage{booktabs, caption, makecell}
\usepackage{multirow}
\usepackage{booktabs}
\usepackage{colortbl}
\usepackage{array}
\usepackage{makecell}
\usepackage{booktabs}
\usepackage{amssymb}
\usepackage{threeparttable}
\usepackage{comment}
\usepackage{authblk}
\title{LOREN: Low-Rank-Based Code-Rate Adaptation in Neural Receivers}

\author{Bram van Bolderik}
\author{Vlado Menkovski} 
\author{Sonia Heemstra de Groot} 
\author{Manil Dev Gomony}

\affil{Eindhoven University of Technology, The Netherlands}

\date{July 2025}

\begin{document}

\maketitle

\begin{abstract}

Neural network based receivers have recently demonstrated superior system-level performance compared to traditional receivers. However, their practicality is limited by high memory and power requirements, as separate weight sets must be stored for each code rate. To address this challenge, we propose LOREN, a Low Rank-Based Code-Rate Adaptation Neural Receiver that achieves adaptability with minimal overhead. LOREN integrates lightweight low rank adaptation adapters (LOREN adapters) into convolutional layers, freezing a shared base network while training only small adapters per code rate. An end-to-end training framework over 3GPP CDL channels ensures robustness across realistic wireless environments. LOREN achieves comparable or superior performance relative to fully retrained base neural receivers. The hardware implementation of LOREN in 22nm technology shows more than 65\% savings in silicon area and up to 15\% power reduction when supporting three code rates.

\end{abstract}

\section{Introduction}


Future wireless technologies, such as 6G, aim to significantly increase throughput, with projections indicating a tenfold improvement per generation (e.g., 5G: 1 Gbps, 6G: 10 Gbps), while simultaneously demanding a tenfold decrease in energy usage per bit~\cite{6gpromise}. This combination of goals heavily constrains the power budget for base-band processing hardware, necessitating the development of innovative, energy-efficient algorithms and system designs. Recently, neural network based receivers (neural receivers) have shown to provide better system-level performance (Block Error Rate (BLER)) performance than traditional receivers across various channel conditions~\cite{Sionna,Honkala2021}. We consider these neural receivers as base neural receivers. Although base neural receivers can achieve strong performance, their high computational complexity has limited adoption in real systems. Several studies have reduced complexity \cite{Bram,BramMEAN}, but the parameter count can still imply large on-chip memory requirements \cite{Meanjourn}. In addition, previous base neural receivers typically assume a fixed CR during training and operation. 
Receivers in mobile communications have a changing code rate (CR), which is the ratio of useful information bits to the total transmitted bits, indicating how much redundancy is added for error correction at the receiver. In base neural receivers, for each CR the entire set of weights needs to be retrained and stored to get good accuracy, which results in a significant amount of extra overhead because of the high parameter count of the network.

\begin{figure}[ht]
  \centering
  \includegraphics[width=0.8\linewidth,trim={0mm 45mm 150mm 90mm},clip]{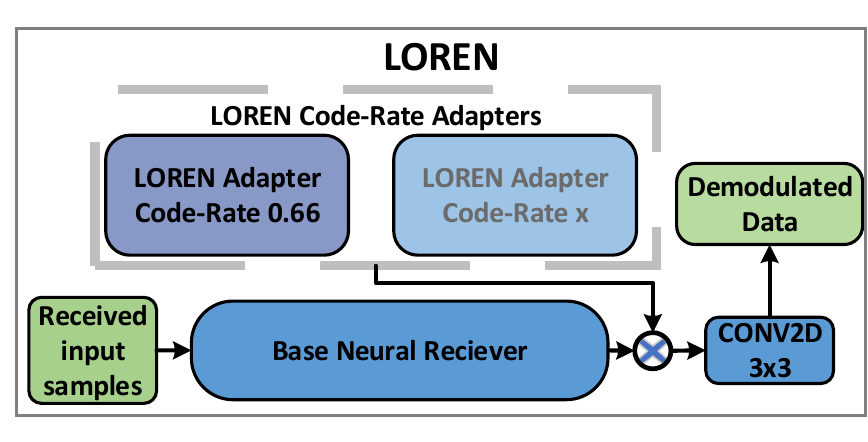}
  \caption{Overview of LOREN architecture that can dynamically adapt to varying CR with the help of layers that are augmented with LOREN adapters for each CR.}
  \label{fig:LORENsimple}
\end{figure}

Low-rank adaptation (LoRA) was recently introduced in large language models as a means to facilitate task specific fine-tuning by learning only a small, low rank update to frozen pre-trained weights~\cite{hu2021loralowrankadaptationlarge}. In this paper, we were inspired by the concept of LoRA, we developed a \emph{code-rate–adaptive} system that requires only a small number of parameters for each CR, yet still delivers performance on par with or better than full weight retraining. We designate this architecture as \emph{LOREN (low rank-based code-rate adaptation neural receiver)}. Figure~\ref{fig:LORENsimple} gives an overview of the LOREN architecture. The base neural receiver consists of a convolutional neural network (NN) based on Sionna neural receiver for OFDM SIMO Systems ~\cite{Sionna} architecture. However, LOREN is trained with input data from a combination of CR, after which the weights in the base neural receiver are frozen. The base neural receiver receives input samples for demodulation. However, this alone will not provide an acceptable accuracy compared to having a network trained specifically for one CR. Therefore, at run-time, the receiver selects an appropriate LoRA adapter(trained for a specific CR) based on the code rate. This LOREN adapter is significantly smaller than an entire weight set in a base neural receiver. This adapter fine-tunes the output of the network for a specific CR which drastically reduces memory footprint and power usage compared to base neural receivers, which use separate weight sets for each CR.

The main contributions of this paper are as follows. First, we introduce LOREN, a novel low-rank-based code-rate adaptation neural receiver that integrates novel LOREN adapters into convolutional layers, enabling dynamic switching between multiple CR with minimal parameter overhead. Second, we present an end-to-end training procedure with per-CR adapters, where the base network weights are frozen and only the small $(A,B)$ matrices are learned for each CR using a standard Single Input, Multiple Output (SIMO) link simulation. Third, we provide a comprehensive performance evaluation by simulating a SIMO system under varying channel conditions, demonstrating that LOREN matches or outperforms fully retrained base neural receivers in BLER. Finally, we conduct a hardware cost analysis by synthesizing both classical and LOREN receivers in 22nm FD-SOI technology using an automated High-Level Synthesis (HLS) flow, showing that LOREN reduces total power by up to 15\% and silicon area by over 65\% when supporting three CR.

\section{Related Work}
\label{relatedwork}

Recent studies indicate that NN-based receivers exhibit better BLER performance compared to conventional models under various channel conditions~\cite{Sionna,Honkala2021}. However, these neural receiver models have a substantial number of parameters, leading to considerable power consumption and large area consumption. Consequently, methods or network architectures that mitigate these factors are vital. One promising approach is the use of Dynamic Neural Networks (DyNNs) which allow runtime adaptability through mechanisms such as early exiting, layer skipping, or depth control~\cite{Han2021}. For example, a layer skipping method proposed by~\cite{Bram} involves the use of a gating network to bypass convolutional blocks. Such principles have been applied in wireless receivers by implementing Signal to noise ratio (SNR) based layer skipping to conserve power. Despite their effectiveness, these techniques typically require additional weight sets for each skipping configuration, as well as CR or additional control logic.

Mixture-of-Experts (MoE) engages specific sub networks tailored to particular inputs. ~\cite{ShazeerMMDLHD17} introduced sparse MoEs for natural language processing applications, and HydraNet~\cite{8578941} expanded this approach to convolutional neural networks (CNN). In the realm of wireless communications, MEAN~\cite{BramMEAN,Meanjourn}  employs expert CNNs chosen based on channel quality. However, adapting such systems to a receiver capable of handling multiple CR presents challenges. This difficulty arises from the necessity of having different weight sets, which in turn increases memory and power usage, a notable disadvantage compared to conventional receivers prevalent in the industry.
LoRA~\cite{hu2021loralowrankadaptationlarge} provides an efficient approach by introducing small low-rank updates to frozen weights to natural language processing. LoRA has been applied to convolutional scenarios for image enhancement~\cite{chai2025}. More recent developments in LoRA include DyLoRA~\cite{DyLoRA2023}, which enables dynamically adjusting the rank during training, as well as analytical studies such as ~\cite{Zeng2024} that seek to quantify LoRA’s representational power. Nevertheless, previous LoRA work primarily focused on fixed fine-tuning performed during training rather than on adaptive changes at runtime, applying it during inference and targeting convolutional networks, which can still benefit from LoRA-inspired methods.

\section{Background}

\label{background}

\begin{figure}[ht!]
\centering
\includegraphics[width=0.8\linewidth,trim={290mm 160mm 10mm 65mm},clip]{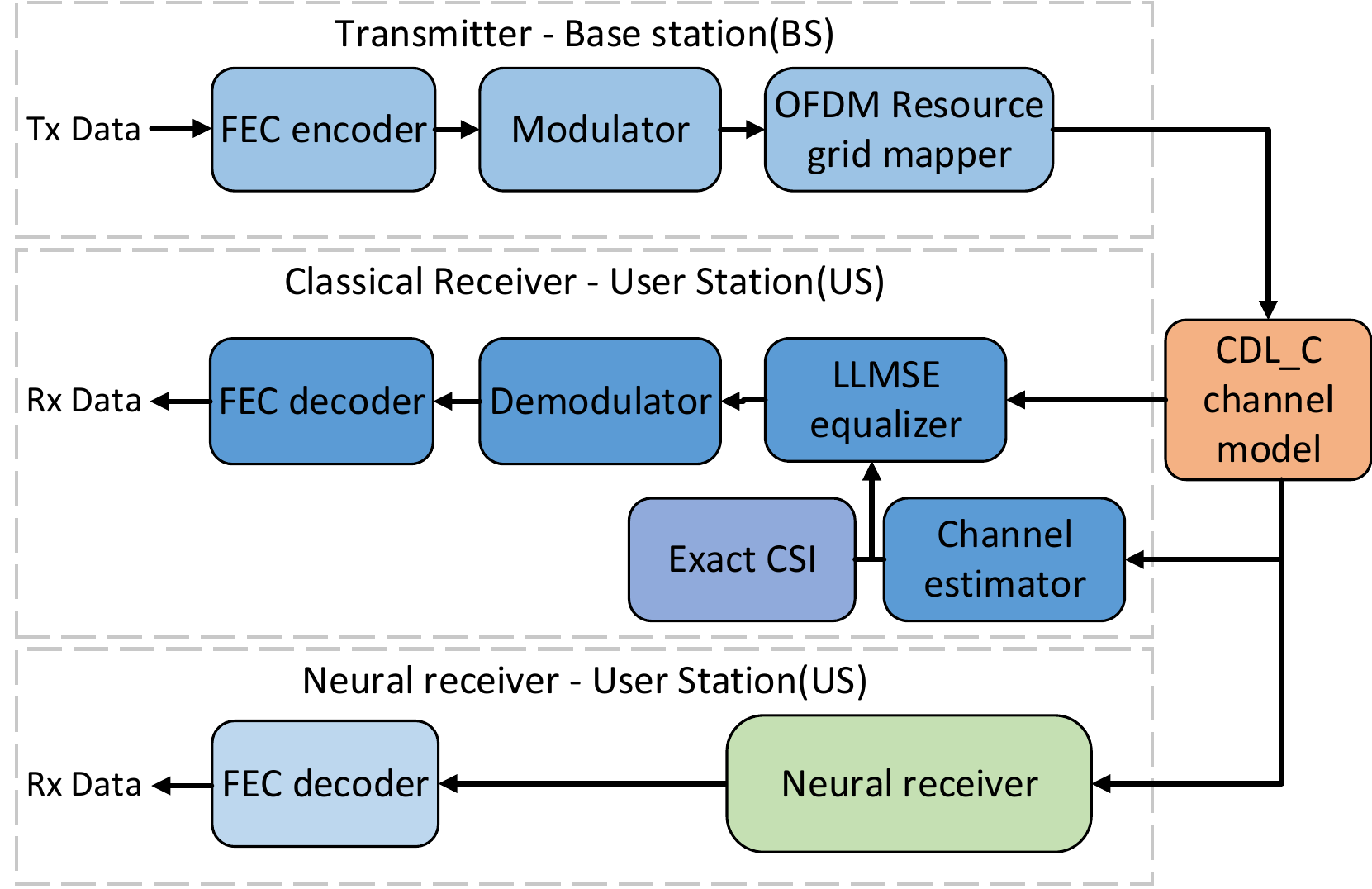}
\caption{Overview of SIMO systems featuring a comparison between a classical receiver baseline, which employs a channel estimator with an option to use either exact CSI for an ideal theoretical performance or employing Least Squares (LS) estimation, and a neural receiver\cite{Sionna}, analyzed using a CDL\_C channel model.}
\label{fig:overview}
   
\end{figure}

 The LOREN framework is built on \emph{Sionna} \cite{Sionna}, which is state of the art in neural receivers, adopting a similar ResNet-style architecture. Although we applied this technique to a residual-style NN architecture, it is also applicable to other convolutional NN architectures. An overview of the SIMO system is depicted in Figure ~\ref{fig:overview}. The schematic representation of this communication framework comprises the following components: The transmitter at the Base Station (BS) includes a Forward Error Correction (FEC) encoder, a (de)modulator, and an Orthogonal Frequency Division Multiplexing (OFDM) mapper. The FEC encoder utilizes a Low Density Parity Check (LDPC) code for encoding to enhance data quality and reliability in noisy environments. Two types of classical receivers from industry are used as benchmarks. The classical receiver utilizes the Linear Least Mean Squared (LS) algorithm, which uses the Channel State Information (CSI) to mitigate noise effects within the communication channel. This CSI includes data on the impact of the channel on signal quality, incorporating aspects such as scattering, fading, and power attenuation over distance. The receiver can utilize ideal channel information for an ideal performance benchmark or the industry standard LLMSE~\cite{Honkala2021}. The classical receiver model is based on the architecture of the Sionna neural receiver for OFDM SIMO systems~\cite{Sionna}.

\section{Proposed LOREN Architecture}
\label{sec:LORAarch}
\begin{figure}[ht]
  \centering
  \includegraphics[width=0.9\linewidth,trim={235mm 10mm 20mm 35mm},clip]{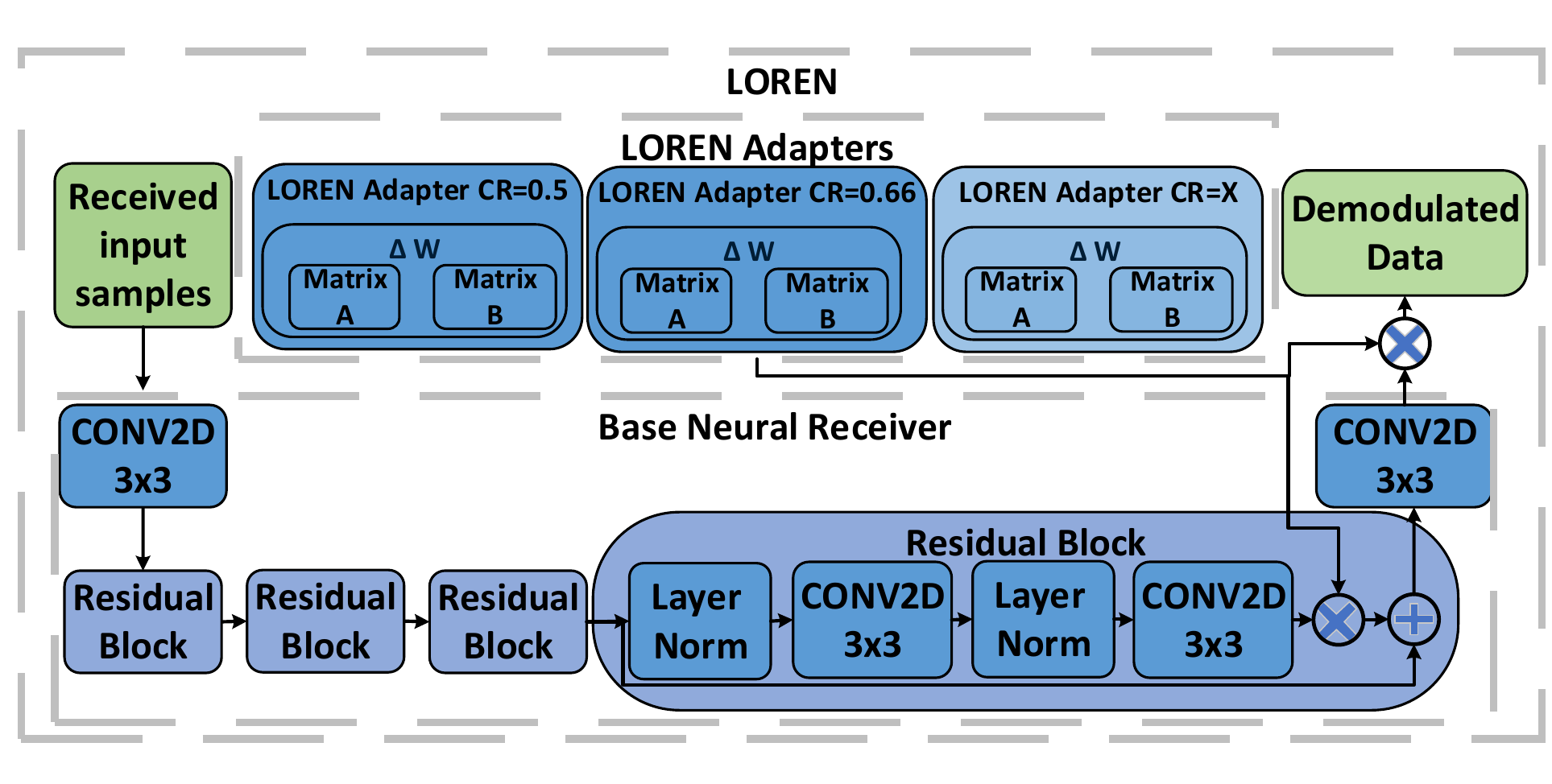}
  \caption{Overview of LOREN two-layer configuration that can dynamically adapt to varying Code rates(CR) with the help of layers that are augmented with per-CR LOREN adapters.}
  \label{fig:LOREN}

\end{figure}

This section introduces LOREN, a neural receiver that facilitates adaptable CR switching through LOREN adapters. A detailed overview of the LOREN architecture is shown in Figure \ref{fig:LOREN}.

\subsection{Base neural receiver architecture}
\label{archrec}



Figure~\ref{fig:LOREN} shows the LOREN architecture. The base neural receiver is identical except that it omits the LOREN adapters. It comprises input/output convolutional layers and four residual blocks, each with two layer-normalization layers and two convolutional layers connected via a skip connection, totaling about 4.8 million parameters (weights and biases) per coding rate/modulation scheme. Training data is generated by sending randomized packets through the transmitter and channel model in Fig.~\ref{fig:overview} across the full noise range by varying channel conditions, but despite outperforming traditional receivers, the base model is complex and parameter-heavy and requires separate weight sets for different modulation schemes and CR.

\subsection{LOREN}

LOREN is inspired by LoRA. LoRA is used where large-scale pre-trained Transformers are expensive to fully fine-tune, so LoRA freezes the original weights $W_{0}$ and learns a low-rank update $\Delta W=\frac{\alpha}{r}BA$, reducing training time~\cite{hu2021loralowrankadaptationlarge}.

Figure \ref{fig:LOREN} shows that each residual block is composed of two CONV2D layers and two layer-normalization layers, incorporating a skip connection for enhanced performance. In contrast to base neural receivers, this set-up includes LOREN adapters for two layers tailored for different CR. 

Base neural receivers need each convolutional layer, characterized by a kernel size of \(3\times3\) and channel dimensions input channels (\(C_{\rm in})= \) output channels (\(C_{\rm out})=128\), to allocate \(128 \times 128 \times 3 \times 3 = 147456\) parameters per CR. Similarly, each layer-normalization layer demands \(128\times128\times14\) parameters. Given that there are eight CONV2D layers along with eight layer-normalization layers, maintaining separate weight sets for various CR results in a large on-chip memory.

In LOREN, we propose that certain CONV2D layers are substituted by CONV2D layers with LOREN adapters applied to them LORENCONV2D layers, where the base weights of the network \(W_0\in\mathbb{R}^{3\times3\times128\times128}\) remain frozen, and for each of the three CR, a low rank adapter is trained. These adapters fine-tune the base network for a specific CR. Employing these low rank adapters removes the need to store separate complete CR weight sets on chip memory, greatly reducing the impact on hardware. For example, instead of \(147456\times3=442368\) parameters per CONV2D layer in a base neural receiver, we only need \((128+128)\times4\times3=3072\) parameters with LOREN to adapt to three CRs.

LOREN employs LOREN adapters which are different to traditional LoRA in that that is used to adapt LLM's during training, whereas LOREN is based on a convolutional network and uses its adapters to adapt during inference.
Unlike traditional LoRA, which modifies the full spatial filter, LOREN restricts the low-rank update to a lightweight $1\times1$ convolution applied independently at each $(t,f)$ location in the time–frequency grid of the received OFDM signal. The input of each convolutional layer has the shape \(\text{Batch} \times T \times F \times C\), where $T$ is the number of OFDM symbols, $F$ the number of subcarriers and $C$ the number of feature channels; each cell $(t,f)$ contains a feature vector of size $C$. The low-rank adapter performs a small channel-wise projection: $A^{(\mathrm{CR})}$ compresses the $C$-dimensional vector to size $r$, and $B^{(\mathrm{CR})}$ projects it back to $C$. This operation focuses purely on mixing feature channels at each location, leaving the expensive spatial filtering in $W_{0}$ intact. The key intuition is that channel correlations across time and frequency are largely stable and already captured by frozen $3\times3$ kernels of the base network, while CR mainly affects how feature channels are combined, which is efficiently handled by the $1\times1$ update. As a result, only $C_{\rm in}\times r + r\times C_{\rm out}$ new parameters are added per CR, drastically reducing memory requirements and enabling instant switching between CRs without retraining the entire network.

\subsection{Training of the base network}
\label{sec:BaseTraining}

The base neural receiver (consisting of all \(W_{0}\) kernels) is trained end-to-end over the entire SNR spectrum by uniformly sampling noise variances and sending random data packets through the 3GPP CDL channel and transmitter (as shown in Fig.~\ref{fig:overview}). This base network is trained with the randomized CR to provide a general base that the LOREN adapters can fine-tune.  The NN is optimized by back-propagating through the transmitter, channel, and baseline neural receiver. This results in a set of weights for randomized CR for the base neural receiver.

\subsection{Training of the LOREN adapters}
\label{sec:LORATraining}

In contrast to traditional LoRA~\cite{hu2021loralowrankadaptationlarge}, which updates every entry of dense or attention
weight matrices and merges the low rank update $\Delta W$ back into $W_0$ for inference, LOREN adapts
the concept to convolutional neural receivers with per-CR fine-tuning. Specifically, the
$3\times3$ base kernels $W_0$ are frozen and only two small matrices are learned for each CR:
\[
A^{(\mathrm{CR})}\in\mathbb{R}^{C_{\rm in}\times r},\quad
B^{(\mathrm{CR})}\in\mathbb{R}^{r\times C_{\rm out}},
\]
\[
W^{(\mathrm{CR})} = W_0 + \frac{\alpha}{r} A^{(\mathrm{CR})} B^{(\mathrm{CR})}.
\]

 During each gradient update, a noise variance and a random CR are selected from a specified list of target CR. For the purpose of this experiment [0.5, 0.66, 0.75] are shown, but practically any CR can be used. Subsequently, OFDM symbols are generated and transmitted through a 3GPP CDL channel, and the output from the NN is back-propagated via a LOREN. This method of using per-CR adapters decreases the number of trainable parameters from \(O(K^2C_{\rm in}C_{\rm out})\) down to \(O\bigl(r\,(C_{\rm in}+C_{\rm out})\bigr)\) for each CR. Here, \(r\) represents the rank and \(k\) the kernel size. It also allows for instantaneous transitions between error-correction configurations without retraining the necessitating full network. These features, i.e., spatially localized adaptation \(1\times1\) and dynamic per-CR expert selection constitute the novel contributions of LOREN relative to conventional LoRA in language models.

\section{Performance evaluation of LOREN}

\label{performance}

For the purpose of performance analysis, we use the wireless system model of a SIMO system, as introduced in Section~\ref{background}, with the following parameters: FFT size of 128, sub-carrier spacing of 30~kHz, 14~OFDM symbols, and carrier frequency of 3.5~GHz. We used a QAM-16 modulation scheme and a CR set of [0.5, 0.66, 0.75]. It should be noted that the NN model was trained using a CDL-C channel model. In this study, we used these three CR as examples to show a sufficient variation in CR. However, retraining enables the network to support any desired CR values. By adding or removing LOREN adapters, any number of CR configurations can be supported (e.g., switching between 2, 3, 4 or more CR).

\begin{figure}[h!]
  \centering
  \includegraphics[width=0.9\linewidth,trim={0mm 0mm 0mm 0mm},clip]{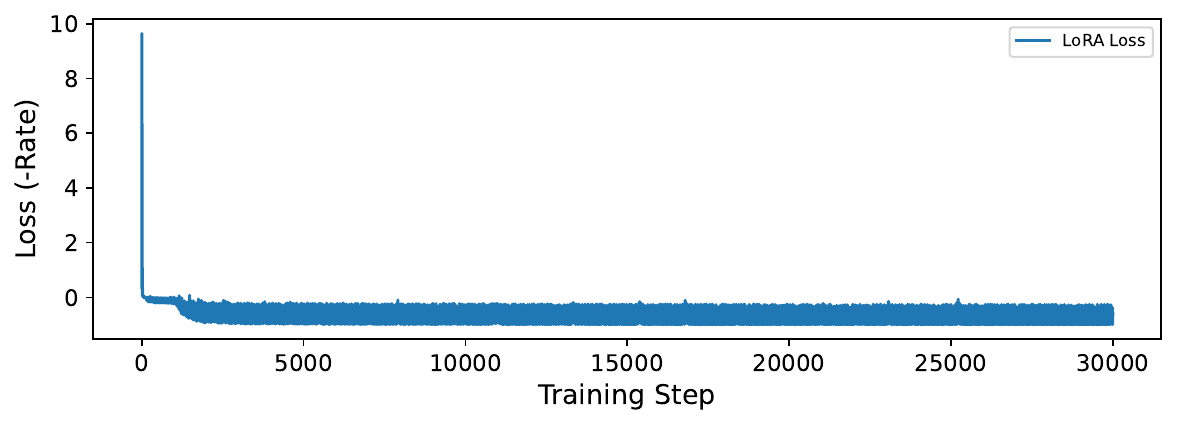}
  \caption{Loss plot for LOREN of QAM16, showing the changes in the loss value over the training iterations. }

   \label{fig:LoreLoss}
     
\end{figure}

Figure \ref{fig:LoreLoss} illustrates the loss metrics for training of LOREN. Initially, the loss values appear elevated, but approximately 30,000 iterations are required to achieve good performance. Each iteration corresponds to a Stochastic Gradient Descent (SGD) training step. This constitutes a single iteration in training, where the gradient is computed and applied to update the weights.

\begin{figure}[h!]
  \centering
  \includegraphics[width=1\linewidth,trim={0mm 0mm 0mm 0mm},clip]{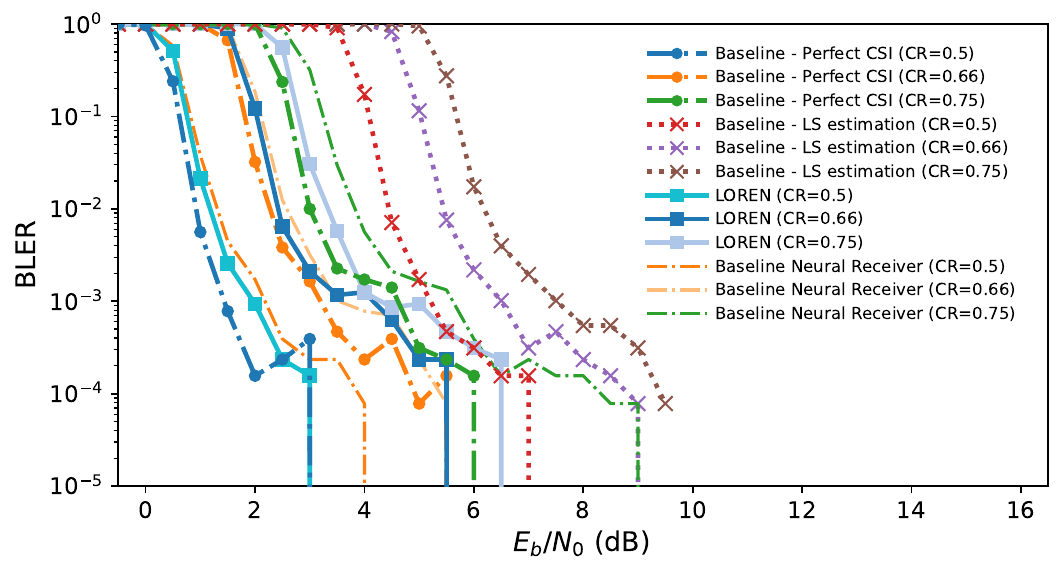}
  \caption{BLER plot depicting the performance of LOREN vs classical approaches and baselines. This shows that LOREN matches or outperforms classical approaches as well as least squares estimation.}

   \label{fig:LoreBasicov}
\end{figure}

Figure~\ref{fig:LoreBasicov} presents the BLER performance of the proposed LOREN network evaluated at three different CRs, using two LORENCONV2D layers, a rank of 4, and an alpha value of 1. Additional architectural configurations are examined in the following subsection. The figure also includes two benchmark receivers: (i) Baseline – Perfect CSI, which represents the theoretical upper bound on performance, and (ii) Baseline–LS Estimation, whose results are shown only for a CR of 0.66 to preserve figure readability. The Baseline – LS Estimation corresponds to a conventional non-neural receiver. In addition, base neural receivers (described in Section~\ref{background}) are trained and evaluated independently for each CR. The block error rate (BLER) is defined as the ratio of incorrectly decoded blocks, each comprising multiple data samples, to the total number of transmitted blocks. The horizontal axis denotes the signal-to-noise ratio, expressed as the energy per bit to noise power spectral density ratio, denoted by Eb/N0.

As illustrated in the figure, for a CR of 0.66, LOREN achieves performance comparable to, and in some cases exceeding, both the base neural receiver and the industry-standard LS estimation baseline. Moreover, its performance closely approaches that of the Baseline–Perfect CSI. For other CR values, LOREN consistently outperforms traditional receiver implementations. Base neural receivers were also evaluated at CRs for which they were not explicitly trained; however, these results exhibited a noticeable performance degradation and were omitted from the figure for clarity. At first glance, outperforming the base neural receiver may seem counterintuitive, as training a dedicated set of weights for a specific CR would seemingly offer greater representational capacity. A plausible explanation is that optimizing a full weight set for a single CR increases the risk of overfitting to that specific operating point. In contrast, LOREN employs a shared base network across multiple CRs, resulting in a more generalizable feature extractor that is subsequently adapted via lightweight LOREN adapters. It is also worth noting that further hyperparameter optimization could potentially improve the performance of the base neural receiver; however, such tuning is considered beyond the scope of this work. Despite both approaches using comparable training time, these factors suggest that LOREN may be easier to train and more robust across varying CRs.

Note that the performance of LOREN must be interpreted within the context of each CR. As the CR increases, the bit error rate naturally degrades, so the results across different CR are not directly comparable. For example, while LOREN with $\mathrm{CR}=0.5$ may appear most effective in the BLER plot, it does not imply that it outperforms higher CR; each curve should only be compared to other results evaluated at the same CR.

\begin{figure}[h!]
  \centering
  \includegraphics[width=\linewidth,trim={0mm 0mm 0mm 0mm},clip]{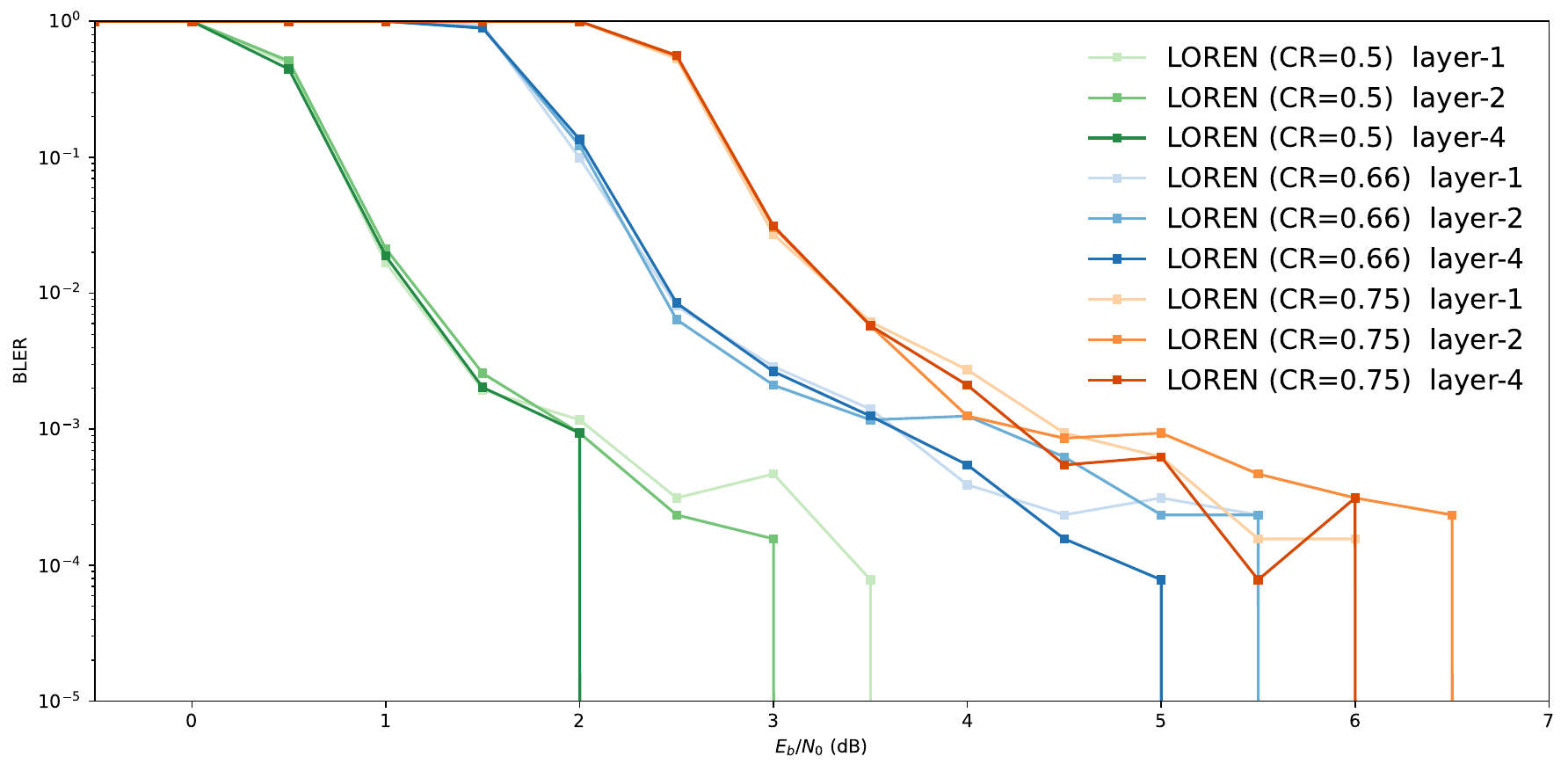}
  \caption{Sensitivity study of impact of LORACONV2D layers. The figure depicts a BLER plot showing the performance of several separate LOREN architectures with different number of LORACONV2D layers applied.}

  \label{fig:LORALAyer}

\end{figure}

Figure \ref{fig:LORALAyer} shows the evaluation of three variants of LOREN analyzed at three different CR. Each variant incorporates 1, 2, or 4 layers of LORENCONV2D. The figure illustrates that an increase in LORENCONV2D leads to improved receiver performance. As in each case, the 4 layer variant outperforms the other implementations. Therefore, for better performance, a higher number of layers using LOREN adapters is beneficial. However, this advantage comes at a cost as additional \(A\) and \(B\) matrices are required for each added layer. Although this extra requirement is substantially smaller than the overhead of a full set of weights, it still contributes to the total system overhead, which is discussed in more detail in Section \ref{hardwaresynth}.

\subsection{Impact of LOREN rank and alpha}

\begin{figure*}[h!]
 \centering
 \includegraphics[width=\textwidth,trim={0mm 0mm 0mm 0mm},clip]{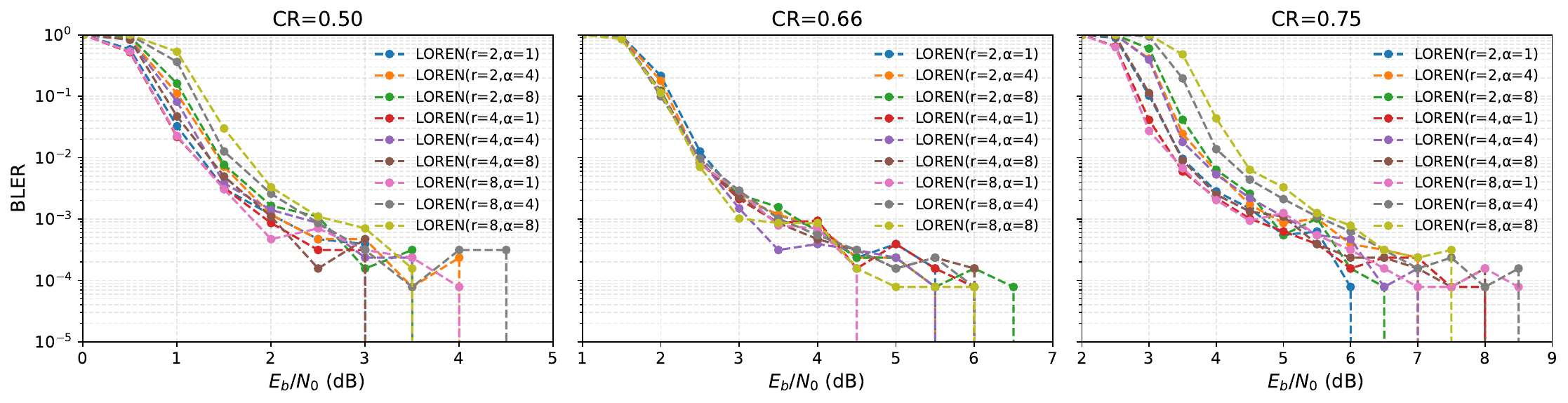}

 \caption{Sensitivity study of LOREN adapter hyperparameters. BLER plots for each CR comparing different ranks ($r \in \{2,4,8\}$) and scaling factors ($\alpha \in \{1,4,8\}$). }

 \label{fig:LORAalpharank}
 \vspace{-6mm}
\end{figure*}

To assess the sensitivity of LOREN to adapter hyperparameters, we performed a joint sweep of rank ($r \in \{2,4,8\}$) and scaling factor ($\alpha \in \{1,4,8\}$) for each CR, as shown in Figure~\ref{fig:LORAalpharank}. For clarity, the classical receivers and baseline references are omitted from these plots, as we have shown in Section \ref{performance} that LOREN outperforms the classical receivers and the baseline. 

The results show that for CR = 0.50 and CR = 0.75, the lower ranks ($r=2$ and $r=4$) outperform $r=8$, despite the latter offering more trainable parameters. This somewhat counterintuitive result can be explained by the scaling factor $\alpha/r$: as rank increases, the scaling factor decreases, which can diminish the effect of additional parameters. In contrast, for CR = 0.66, $r=8$ exceeds the lower ranks. 

In all scenarios, $\alpha=1$ consistently yields superior results, suggesting that larger scaling factors tend to destabilize low rank updates rather than improve them. Importantly, these findings highlight that performance does not improve monotonically with rank or $\alpha$, and instead depends on a balance between expressivity and stability.   

LOREN has a trade-off in preserving general knowledge in the frozen backbone and enabling sufficient task/condition-specific adaptation through the adapters. In LOREN these are specifically controlled by the adapter rank $r$,  the scaling factor $\alpha/r$ (update magnitude), and the LoRA adapter count  (only the last layer vs more layers). Note that these rank and alpha choices are primarily empirical. An extensive parameter sweep (e.g., exploring additional ranks beyond 8 or finer $\alpha$ values) could provide further insight into the interaction between rank, scaling, and CR. We leave such an expanded sensitivity study to future work as it was outside the scope of this study.

\section{Hardware implementation}
\label{hardwaresynth}

In this section, we perform a hardware complexity analysis by implementing a base neural receiver and LOREN in hardware. Note that a direct hardware complexity comparison with classical non-neural receivers is not provided, as published results include additional system components, and synthesizing a comparable baseline was beyond the scope of this work. To assess the area and power characteristics of LOREN, the architecture is first implemented at the Register Transfer Level (RTL) level and then subjected to logic synthesis using an automated workflow based on earlier work\cite{Bram}. This approach enables an accurate comparison of power consumption and silicon area utilization between LOREN and base neural receivers.

Initially, a classical receiver or LOREN model is trained in TensorFlow, undergoing fine-tuning until it achieves the desired performance metrics. SystemC code is then generated from layer-specific templates (including LORACONV2D, CONV2D, and layernorm). The HLS tool, Cadence Stratus \cite{Stratus}, is then utilized in conjunction with these SystemC models and synthesis configurations to produce an RTL level design. This design is further processed to derive precise power and area metrics. Both the base neural receiver and the LOREN architecture were synthesized using this approach in 22nm FD-SOI technology, operating at 200 MHz@0.9V, using fast latency constraints as well as a high optimization effort. A uniform 16-bit quantization was implemented to prevent accuracy degradation due to quantization in the hardware implementation.

Evaluating the hardware cost of LOREN requires accounting for the SRAM used to store the shared base network weights, the per-CR weight sets of the classical receiver, and the additional LOREN adapter weights. As shown in Table \ref{tab:memory}, the memory demands for storing the weights of each layer significantly exceed those for the LOREN adapters. This is particularly notable given the presence of 8 layer-normalization and 8 CONV2D layers. The layers require multiple SRAMs connected in parallel per layer to store the weights. For a LOREN configuration of a given rank, only one of the three LOREN SRAMs is required. The specific LOREN SRAM used is determined by the particular LOREN configuration.

In neural receivers, changing weight sets causes additional latency as a result of the time required to access the weights. Specifically, in the context of our mobile communication application, as per the 3GPP 5G NR standard, the subframe length is approximately 1 \(ms\) (which is the minimum time window during which we could assume no need for change) if we account for four SRAMs of 4K each summing up to 16k, this matches the approximate number of trainable weights in a convolutional layer with dimensions of 144 (obtained from a \(3\times3\) kernel and 16-bit resolution), thus necessitating one set per convolutional layer. In total, each convolutional layer requires a set. Consequently, if all memories were accessed concurrently, a layer could be read in 4096 $\times$ 5ns = 20.48 \(\mu s\). Within the entire subframe, this allows approximately 48 reads to be performed within a single subframe. For layer-normalization, it takes the same amount of time, and for the LOREN adapters it is a lot faster because those memories are considerably smaller, proving that switching is possible within each subframe. The hardware results detailed below account for power consumption for reading out weights as a result of accessing LOREN adapters or, in the classical receiver's case, extra CR sets in base neural receivers.

Other techniques to be more parameter efficient such as fine-tuning even a few Conv2D layers still introduce a lot of parameters (and would need a separate set per code rate), while LOREN stays much smaller because it uses low-rank A/B matrices, and can be made smaller still with a simple linear 1×1 mixer.

Bottleneck adapters\cite{chen2024convadapterexploringparameterefficient} differ mainly because they are nonlinear, increasing the risk of overfitting and adding compute/latency. Since changing code rate affects the way  the same received signal is interpreted in a larger way than changing the underlying patterns of the signal, a linear 1×1 mixer that just re-mixes the channel features already-extracted should be sufficient nonlinear adapters are more useful only if changes in code rate create genuinely new pattern of characteristics. A direct comparison within a neural receiver for standard bottleneck adapters and fine-tuning only a small number of Conv2D layers would be valuable, but this is beyond the scope of this work and left for future research.

\begin{table}[h!]
\centering
\setlength{\tabcolsep}{1pt} 
\renewcommand{\arraystretch}{1.0} 
\begin{tabular}{|l|r|r|r|}
\hline
\rowcolor[HTML]{9B9B9B}
\textbf{Operation}
  & \makecell{\textbf{Memory words}}
  & \makecell{\textbf{Memory bitwidth}}
  & \makecell{\textbf{SRAMs layer}} \\
\hline
\rowcolor[HTML]{FFFFFF}
Convolution                      & 4096  & 144 & 4   \\ \hline
\rowcolor[HTML]{C0C0C0}
Convin+Convout                   & 896   & 144 & 1   \\ \hline
\rowcolor[HTML]{FFFFFF}
Layernorm                        & 4096  & 224 & 8   \\ \hline
\rowcolor[HTML]{C0C0C0}
LOREN rank-2-768                  & 768   & 16  & N/A \\ \hline
\rowcolor[HTML]{FFFFFF}
LOREN rank-4-1536                 & 1536  & 16  & N/A \\ \hline
\rowcolor[HTML]{C0C0C0}
LOREN rank-8-3072                 & 3072  & 16  & N/A \\ \hline
\end{tabular}
\caption{Memory requirements and SRAM usage for storage of Base network layer parameters and LOREN adapters.}
\label{tab:memory}

\end{table}

\subsection{Power and area consumption of LOREN}

\begin{figure}[h!]
    \centering
    \includegraphics[width=0.8\linewidth,trim={0mm 0mm 0mm 0mm},clip]{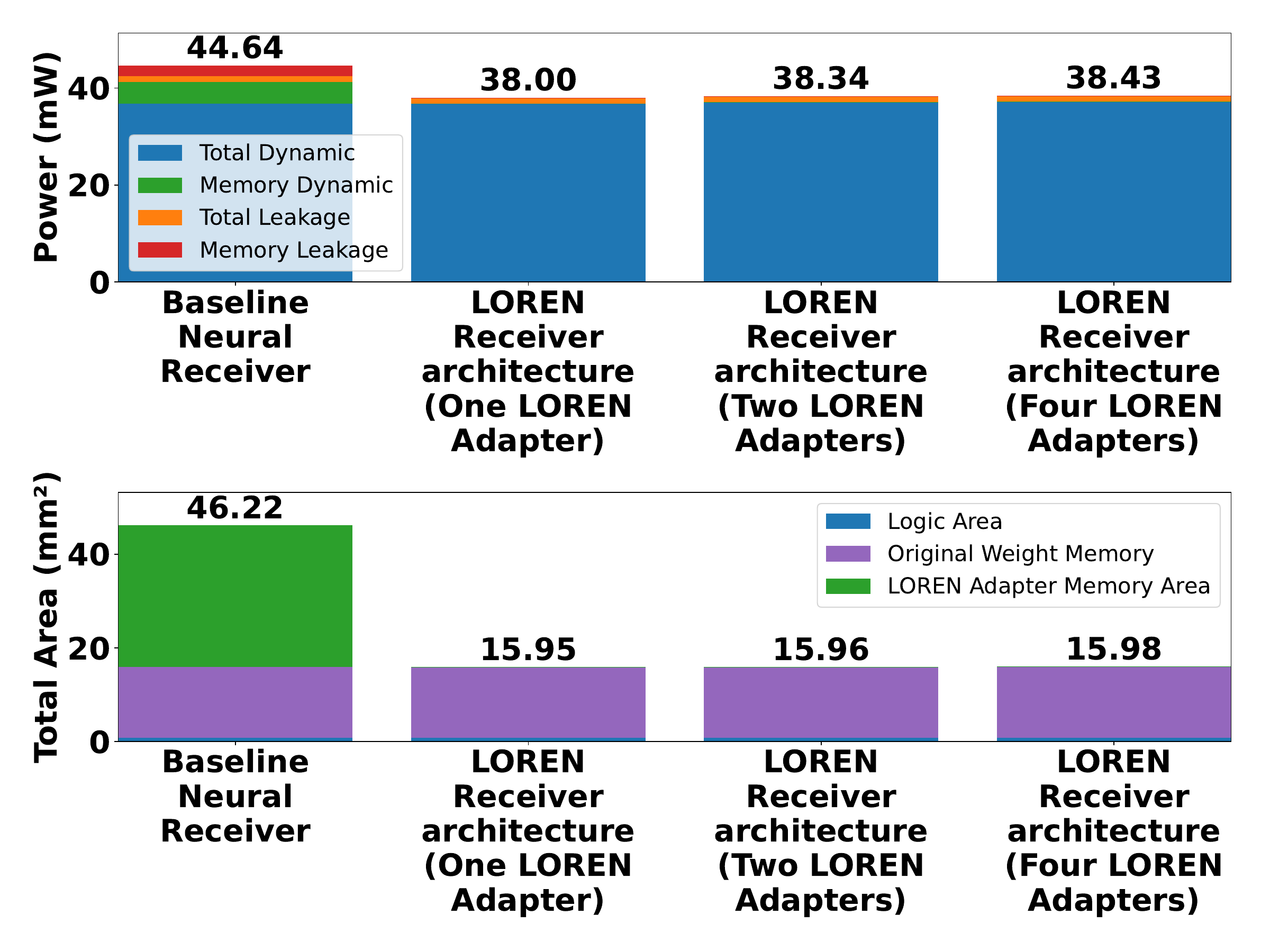}
    \caption{Hardware synthesis results for LOREN in 22 nm FD-SOI technology, operating at 200 MHz , 0.9V. The graph on top illustrates shows power vs baseline. The graph below depicts the same in terms of total area.}

    \label{fig:Loreareapower}
\end{figure}

Figure \ref{fig:Loreareapower} illustrates the results of the hardware synthesis for LOREN. This includes logic, leakage, and dynamic power, along with the leakage and dynamic power of the original weight memory. The dynamic power includes the memory read access power for both the classical implementation and the LOREN model.  Each SRAM was synthesized using the GlobalFoundries memory compiler for the 22nm FDSOI technology (single-port low-power S1PL macro) according to the specifications in Table \ref{tab:memory}; the compiler generated datasheets for the desired SRAM capacities provide the power and latency values used for the calculations in this work.

For the classical receiver, the CR memory holds the weights of each CR, which comprises three weight sets, one for each CR, is also depicted. Power numbers for LOREN implementations with 1, 2, and 4 layers are presented, specifically for logic, leakage, and dynamic power. This includes the memory for the LOREN adapters in the case of LOREN. Evidently, LOREN demands significantly less power, given its elimination of extra memories required for additional CR. The addition of more LOREN layers affects power consumption insignificantly, which has a minor impact on total power usage compared to the basic weight memory and logic.

The graph at the bottom of Figure \ref{fig:Loreareapower} illustrates the area, which shows the large difference between base neural receivers and LOREN. The base neural receiver area is substantial because of the additional CR memories. The corresponding area occupied by the LOREN adapters weight is depicted in green on the graph; however, it is not visible in the graph due to its negligible size relative to the entire memory for the base network weights.

From these graphs, it is clear that the LOREN layers have a significant impact on hardware performance compared to swapping out weight sets. The reduction in power of a classical network compared to LOREN with 4 LOREN layers is a 14.8 \% decrease while the total area is reduced by 65.5\%. The difference between the number of LOREN layers is negligible. Incorporating an additional 4 LOREN layer adapters yields optimal performance with minimal impact, as depicted in Figure \ref{fig:LORALAyer}.



LOREN scales efficiently to many CRs because adding low-rank adapters adds minimal overhead, whereas the area of the base neural receiver grows substantially with each additional weight set. Varying adapter rank has a negligible effect on total memory and logic (Table~\ref{tab:memory}), so the corresponding plot is omitted due to space limits. A possible extension is to support different modulation schemes with LOREN adapters using the same techniques, but the architecture must be adapted because changing the modulation alters the number of output channels of the NN, requiring additional logic, which is outside the scope of this work.

\section{Conclusions and future work}
\label{conclusions}


LOREN delivers major performance and hardware gains by using custom LoRA-inspired layers and a tailored training loop, achieving better system level performance than classical per-CR weight training by sharing weights and adapting via CR-specific LOREN layers, which makes multi CR support scalable and drastically reduces impractical memory needs in hardware-limited settings. Hardware analysis shows that using four LOREN layers reduces power by 14.\% and area by 65.5\% versus a base neural receiver, with benefits growing as more CRs are supported. Future work includes extending adapters to different modulation schemes, implementing standard bottleneck adapters for comparison, and running a more extensive hyperparameter search.

\bibliographystyle{ieeetr}
\bibliography{references.bib}
\end{document}